\documentclass[10pt,twocolumn,letterpaper]{article}

\usepackage{iccv}
\usepackage{times}
\usepackage{epsfig}
\usepackage{graphicx}
\usepackage{amsmath}
\usepackage{amssymb}

\usepackage[pagebackref=true,breaklinks=true,letterpaper=true,colorlinks,bookmarks=false]{hyperref}

\usepackage{graphicx}
\usepackage{microtype}

\makeatletter
\newcommand\footnoteref[1]{\protected@xdef\@thefnmark{\ref{#1}}\@footnotemark}
\makeatother

\newcommand{\shorteq}{%
  \settowidth{\@tempdima}{-}% Width of hyphen
  \resizebox{\@tempdima}{\height}{=}%
}

\usepackage{amssymb,fge}

\usepackage{amsmath, amssymb}
\usepackage{subcaption}
\usepackage[utf8]{inputenc}
\usepackage[T1]{fontenc}
\usepackage[linesnumbered,ruled,vlined]{algorithm2e}
\usepackage{graphicx}
\usepackage{amsfonts}
\usepackage{enumitem}
\usepackage{soul}
\usepackage{hhline}
\usepackage{multirow, makecell}
\usepackage{float}
\usepackage{booktabs}

\usepackage{amsthm}
\usepackage{color}
\usepackage{transparent}
\usepackage{footmisc}
\usepackage{setspace}
\usepackage{textcomp}
\usepackage{mathtools}

\theoremstyle{plain}

\newtheorem{prop}{Postulate}[section]

\mathchardef\mhyphen="2D

\usepackage{algorithm2e}

\iccvfinalcopy % *** Uncomment this line for the final submission

\def\iccvPaperID{2194} % *** Enter the ICCV Paper ID here

\ificcvfinal\fi

\begin{document}

%%%%%%%%% TITLE
\title{Aerial Diffusion: Text Guided Ground-to-Aerial View Translation from a Single Image using Diffusion Models }

\author{Divya Kothandaraman, Tianyi Zhou, Ming Lin, Dinesh Manocha \\ University of Maryland College Park}
%\and

%\maketitle

\twocolumn[{%
\renewcommand\twocolumn[1][]{#1}%
\maketitle
\begin{center}
    \centering
    \captionsetup{type=figure}
    \includegraphics[scale=0.35]{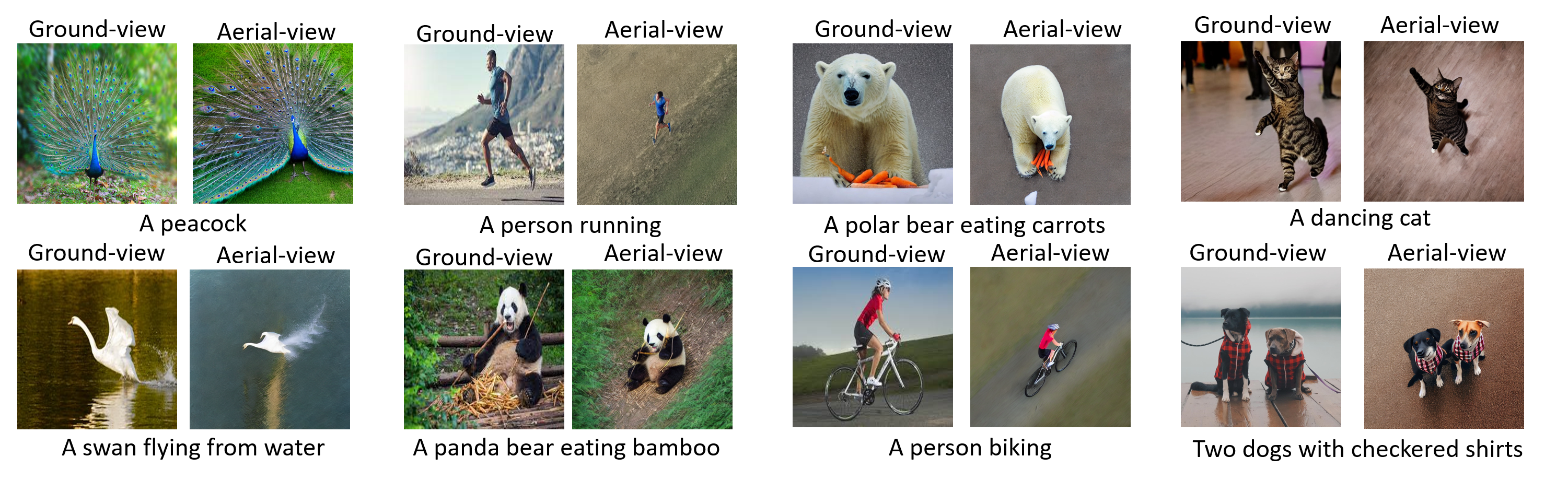}
    \captionof{figure}{\textit{Given a single ground-view image and the corresponding text description as input, Aerial Diffusion generates corresponding aerial-view image.} Our method does not require any supervision from aerial-view data, pairs of ground-aerial view, depth maps, semantic maps, multi-views, etc. It is one of the first approaches to achieve ground-to-aerial view translation in an unsupervised manner.}
    \label{fig:teaser}
    %\vspace{-10pt}
\end{center}%
}]

\ificcvfinal\thispagestyle{empty}\fi

\begin{abstract}
% \vspace*{-1em}
   We present a novel method, {\em Aerial Diffusion}, for generating aerial views from a single ground-view image using text guidance. Aerial Diffusion leverages a pretrained text-image diffusion model for prior knowledge. We address two main challenges corresponding to domain gap between the ground-view and the aerial view and the two views being far apart in the text-image embedding manifold.
   %ground-to-aerial translation is a challenging task. 
   Our approach uses a homography inspired by inverse perspective mapping prior to finetuning the pretrained diffusion model. Additionally, using the text corresponding to the ground-view to finetune the model helps us capture the details in the ground-view image at a relatively low bias towards the ground-view image.  Aerial Diffusion uses an alternating sampling strategy to compute the optimal solution on complex high-dimensional manifold and generate a high-fidelity (w.r.t. ground view) aerial image. We demonstrate the quality and versatility of Aerial Diffusion on a plethora of images from various domains including nature, human actions, indoor scenes, etc. We qualitatively prove the effectiveness of our method with extensive ablations and comparisons. To the best of our knowledge, Aerial Diffusion is the first approach that performs single image ground-to-aerial translation in an unsupervised manner. Code is available at \url{ https://github.com/divyakraman/AerialDiffusion}.
   %i.e. without any paired data, multi-views, depth maps, etc. 
   %We will make all code and data publicly available.
   %\footnote{Code: https://github.com/divyakraman/AerialDiffusion}
\end{abstract}

\section{Introduction}

The analysis of aerial image and video~\cite{kumar2001aerial} plays a pivotal role in different applications, such as search and rescue, aerial photography, surveillance, movie production, etc. However, the paucity of aerial data~\cite{li2021uav} and the complexities associated with data capture from aerial cameras/ UAVs makes it difficult and costly to train large neural networks~\cite{ye2022unsupervised} for these applications. Hence, the synthesis of aerial-view images~\cite{barisic2022sim2air} offers a promising alternative to address these challenges. Conditional image synthesis~\cite{mirza2014conditional,nichol2021glide} allows for control over the generation process. 
% Image synthesis can be either conditional or unconditional. The 

Ground-view annotated datasets~\cite{deng2009imagenet,cordts2016cityscapes,carreira2017quo,monfort2019moments} are readily available for many tasks. Hence, a method that transforms ground-view images to aerial views (or 
cross-view synthesis~\cite{regmi2019cross,tang2019multi}) could allow the reuse the of annotated metadata for a variety of aerial-view applications, e.g., classification~\cite{he2020deep}, segmentation~\cite{hao2020brief}, action recognition~\cite{kothandaraman2022far}, representation learning~\cite{sun2019videobert}, domain adaptation~\cite{choi2020unsupervised}, augmentation~\cite{jaipuria2020deflating}, etc. However, cross-view synthesis requires the network to learn a very large non-trivial translation. The network needs to hallucinate a new and enormously different view of all entities in the scene and the background, while being consistent with the details including the semantics, colors, relations between various parts of the scene, pose, etc. 

Prior work~\cite{regmi2019cross,tang2019multi,ding2020cross,liu2021cross} on ground-to-aerial generation use NeRFs and GANs. However, all of these methods use paired data for ground-view and the corresponding aerial views, which is seldom available. Moreover, training on a specific dataset limits the application to specific scenes similar to the training data; necessitating the collection of paired data for widely different distributions. Instead, our goal is to develop a generic method for generating aerial views from ground-views without any paired data or other auxiliary information such as multi-views, depth, 3D mapping, etc. 

While there are many diverse datasets of ground images, there are not many such good quality aerial datasets~\cite{li2021uav} - hence, unpaired image-to-image translation~\cite{zhu2017unpaired} is not a viable solution. On the contrary, text is an auxiliary modality that can be easily obtained using off-the-shelf image/video captioning tools~\cite{hossain2019comprehensive} %and classification labels~\cite{deng2009imagenet,carreira2017quo} corresponding to ground-view datasets. 
Moreover, text provides a natural representation space describing images. Consequently, our goal is to use the text description of a ground-view image to generate its corresponding aerial view. 

Recently, diffusion models have emerged as state-of-the-art architectures for text-to-image~\cite{kawar2022imagic,hertz2022prompt,zhang2022sine,nichol2021glide} high-quality realistic image synthesis. The availability of immense prior knowledge via large-scale robust pretrained text-to-image models~\cite{rombach2022high}, motivates us to pose ground-to-aerial view translation as text-guided single-image translation~\cite{kawar2022imagic,zhang2022sine}. Text-guided single-image translation methods finetune the diffusion model to the input image and then perform linear interpolation in the text embedding space to generate the desired output. However, direct application of these methods~\cite{nichol2021glide,zhang2022sine,kawar2022imagic} to ground-to-aerial translation either generates high-fidelity non-aerial images or low-fidelity aerial images. \looseness-1

%\begin{figure*}
%    \centering
%    \vspace*{-1em}
%    \includegraphics[scale=0.2]{Figures/results1.png}
%    \vspace*{-1em}
%    \caption{We apply Aerial Diffusion on diverse images such as animals/ birds, natural scenes, human actions, indoor settings, etc and show that our method is able to generate high-quality high-fidelity aerial images.}
%    \label{fig:results1}
%   \vspace{-5pt}
%\end{figure*}

\textit{\textbf{Main contributions.}}
We present two postulates for text-guided image translation, for ground-to-aerial translation: (i) domain gap between the finetuning task (w.r.t. ground view) and target task (aerial view generation) hinders the diffusion model from generating accurate target views and introduces bias towards the source view, (ii) a finetuned diffusion model cannot generalize well to the target prompt, if the text embedding and image spaces corresponding to the source and the target are far apart on the nonlinear text-image manifold.  

\begin{figure*}
    \centering
    \includegraphics[scale=0.2]{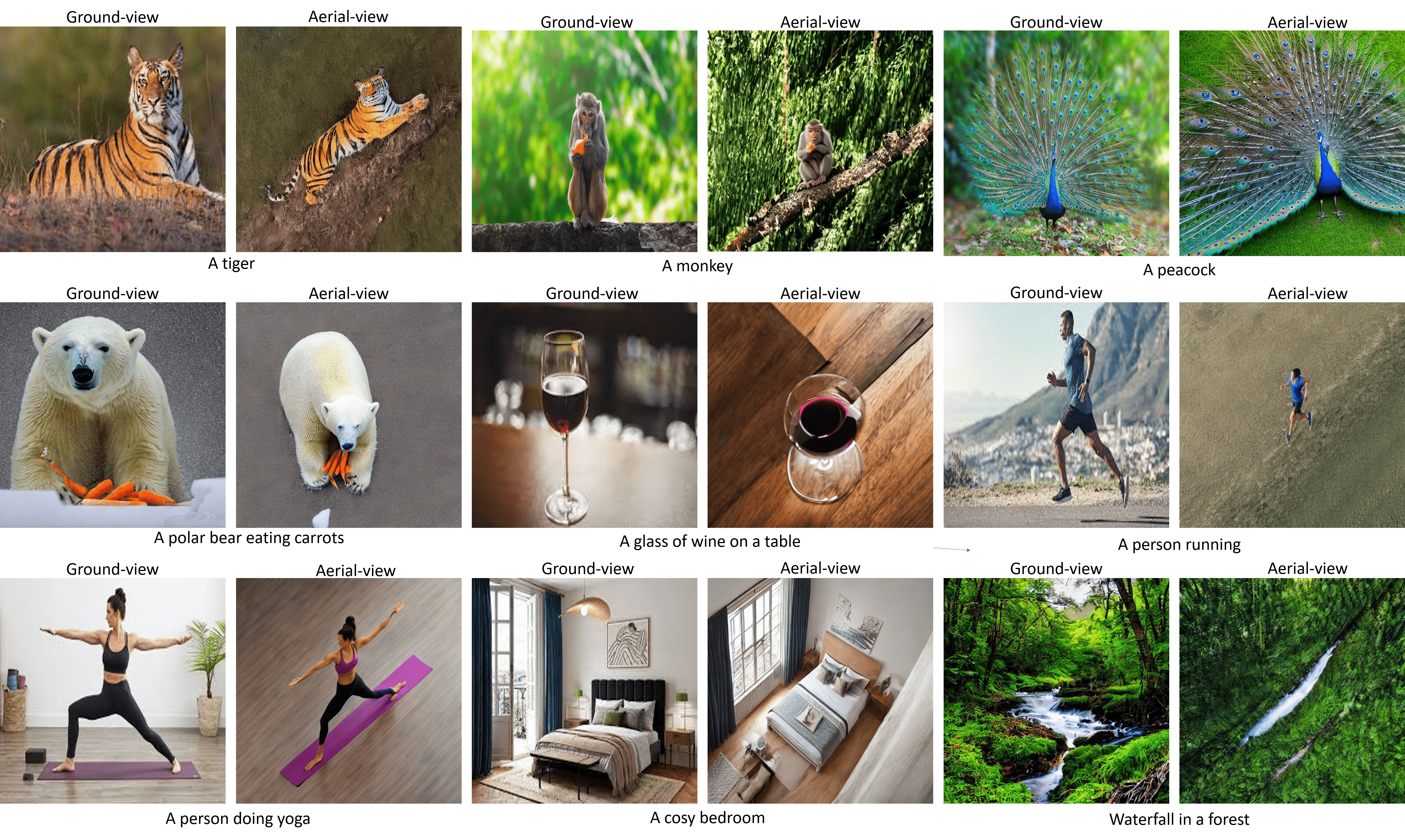}
    \caption{\textbf{Results.} We apply Aerial Diffusion on diverse images such as animals/ birds, natural scenes, human actions, indoor settings, etc and show that our method is able to generate high-quality high-fidelity aerial images.}
    \label{fig:results1}
\end{figure*}

Based on these findings, we propose ``{\em Aerial Diffusion}'', a simple, yet effective, method for generating aerial views, given a single ground-view image and the corresponding text description as input. The novel elements of our algorithm include:
\begin{itemize}[nosep]
    \item Instead of directly finetuning the diffusion model with the ground-view image,  we apply a {\em homography based on inverse perspective mapping} on the ground-view image to obtain a homography projected image prior to the finetuning. This reduces the bias of the diffusion model towards the input image while finetuning.
    \item To finetune the diffusion model, we use the {\em source text} corresponding to the `ground-view' as the {\em guiding factor}, instead of the target text (`aerial view'). This helps the diffusion model search for an optimized text embedding in the vicinity of a text space close to the image space, enabling the learning of a `good' optimized text embedding. This also prevents the diffusion model from developing a bias towards an incorrect aerial view. \looseness-1
    \item To obtain a high-fidelity aerial image (w.r.t. ground-view), at inference time, we manipulate the text embedding layer, such that it prioritizes fidelity and the aerial viewpoint in an alternating manner. {\em Alternating between text embeddings corresponding to the viewpoint and fidelity switches the denoising direction}, such that the backward diffusion takes one step towards preserving fidelity followed by another step towards generating an aerial view. As noises are gradually removed, the process ends up with a high-fidelity aerial-view image on a manifold with a better fidelity-viewpoint trade-off than linear interpolation. %Hence, our mechanism allows the diffusion model to search for the optimal solution in a much higher dimensional space than being restricted to the linear interpolation space.
\end{itemize} 
We apply our method on numerous in-the-wild images from various domains such as nature, animals and birds, human actions, indoor objects, etc. Our method is able to generate high-quality aerial view images that preserve the details contained in the source-view image(Fig.~\ref{fig:results1} and Fig.~\ref{fig:teaser}). We conduct extensive ablation studies (Fig.~\ref{fig:results3}) highlighting the benefits of each element of our method; and demonstrate the trade-off between fidelity (w.r.t. source image) and faithfulness to target view via hyperparameter tuning(Fig.~\ref{fig:results2}). We compare with the state-of-the-art diffusion model-based text-guided editing approach~\cite{kawar2022imagic} and show far superior qualitative performance for ground-to-aerial translation (Fig.~\ref{fig:results3}). Comparison to other text-embedding manipulation approaches (Fig.~\ref{fig:results3}) also shows that our alternate prompting strategy works better. 

\section{Related work}

%Cover image translation, cross view image translation, NeRFs, depth, 3D recon, diffusion models and text based editing, novel view generation

There has been immense work on \textit{image-to-image translation}~\cite{isola2017image,liu2017unsupervised,lin2018conditional,pang2021image,zhu2017unpaired,wang2018video,aldausari2022video} using GANs~\cite{shamsolmoali2021image}, transformers~\cite{esser2021taming}, diffusion models~\cite{dhariwal2021diffusion,rombach2022high,croitoru2022diffusion}, etc. for problems, such as style transfer~\cite{jing2019neural}, image restoration~\cite{su2022survey}, and multimodal style translation~\cite{huang2018multimodal}. Many of these methods are capable of performing these tasks using paired/ unpaired data~\cite{alotaibi2020deep}. Recently, diffusion models~\cite{nichol2021glide,wang2022pretraining,su2022dual,sasaki2021unit,saharia2022palette,saharia2022image,yang2022paint,preechakul2022diffusion} have been successful in performing non-trivial operations, such as posture changes and multiple objects editing. Prior work on \textit{cross-view synthesis}~\cite{regmi2019cross,tang2019multi,ren2022pi,toker2021coming,ding2020cross,ma2022vision,liu2021cross,shi2022geometry,liu2020exocentric,ren2021cascaded,liu2022parallel,wu2022cross,shen2021cross,ammar2019geometric,zhao2022scene} generally use paired data and other complex auxiliary modality such as semantic maps, depth, multi-views, etc within various generative approaches. 

A closely related problem is \textit{novel view synthesis}~\cite{levoy1996light,more2021deep} where the goal is to generate new views of the scene. However, most novel-view synthesis methods including GANs~\cite{xu2019view}, NeRFs~\cite{gao2022nerf}, diffusion models~\cite{watson2022novel} use multiple views of the scene for training, even while they may be capable of performing single-view evaluation~\cite{wiles2020synsin,tucker2020single}. Again, this is prohibitive since it requires multiple views of the scene/ depth information~\cite{hou2021novel} for training. On the other hand, 3D reconstruction methods~\cite{han2019image,yuniarti2019review,fahim2021single} rely on depth information or auxiliary data such as shape priors. Moreover, 3D reconstruction is a complex and expensive task, which is redundant when the goal is to just obtain a 2D aerial view of the scene. 

Text, which is easily available, has been widely used as a guiding factor for image translation~\cite{liu2020describe,li2020image} and image editing~\cite{kawar2022imagic,hertz2022prompt,brooks2022instructpix2pix,kim2022diffusionclip,zhang2022sine,zhuang2021enjoy,hu2021lora,gal2022image,crowson2022vqgan}, particularly in the context of diffusion models recently. The availability of large databases of image-text pairs~\cite{schuhmann2022laion}, open-source pretrained models~\cite{radford2021learning}, and the fact that text is a natural representation of the world makes it conducive to use text as an auxiliary modality to provide guidance. Many \textit{text-based image editing} approaches operate on a single real image~\cite{valevski2022unitune,kawar2022imagic} and perform inference time optimization, making them easy to generalize across diverse images. \textit{Single-image approaches}~\cite{yoo2021sinir,vinker2020deep,ruiz2022dreambooth} have been proposed for image manipulation tasks without text-guidance as well. Generally, they aim to learn useful representations by finetuning a pretrained model on a single image for reconstruction. The inference then controls the feature space~\cite{shen2020interpreting,patashnik2021styleclip} to achieve the desired changes.

\begin{figure*}
    \centering
    \includegraphics[scale=0.20]{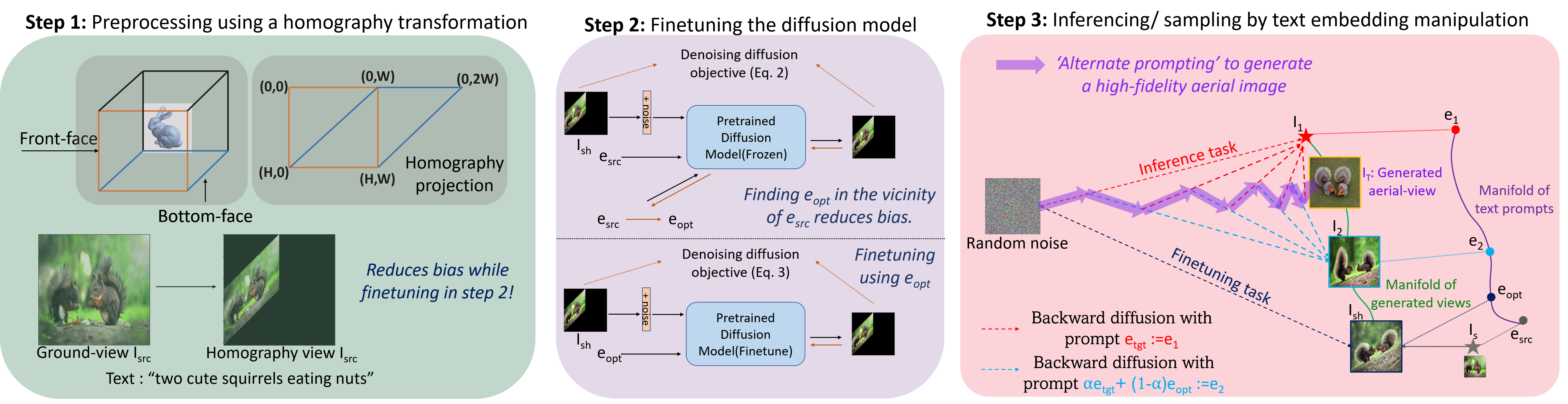}
    \caption{\textbf{Aerial Diffusion.} In step 1, we apply a homography transformation to the ground-view image $I_{S}$. This creates a gap between $I_{Sh}$ and txt, which in turn reduces the bias of the model towards $I_{Sh}$ in step 2. In step 2, we use the $e_{src}$ to optimize to $e_{opt}$ and finetune the diffusion model to reconstruct $I_{Sh}$ for the $e_{opt}$. Using the $e_{src}$ to find $e_{opt}$ reduces the bias of the model towards $I_{Sh}$ due to the disparity between $I_{Sh}$ and the $e_{src}$. In stage 3, we manipulate the text embedding by using an alternating strategy to find the optimal solution in a higher-dimensional non-linear space to generate a high-fidelity aerial image $I_{T}$.
    }
    \label{fig:overview}
\end{figure*}

\section{Aerial Diffusion}

We present Aerial Diffusion for view translation from a single real ground-view source image $I_{S}$ to its aerial-view (or target image $I_{T}$), given a text description txt of the image. The text description can be obtained using an off-the-shelf image captioning tool~\cite{hossain2019comprehensive}. We assume no access to any paired data or other modalities such as depth, semantic maps, other views of the scene, etc. Corresponding to the ground-view, we use the source text description txt$_{G}$ = `front view of' + txt with text embedding $e_{src}$. Similarly, for the aerial-view, we use the target text description txt$_{A}$ = `aerial view of' + txt with text embedding $e_{tgt}$. %We provide an overview of the preliminaries in Section~\ref{sec:prelim}. 
In Section~\ref{sec:theory}, we present two postulates that lead to the method described in Section~\ref{sec:method}.

\subsection{Postulates}
\label{sec:theory}

In this section, we analyze text-guided single image translation in the context of ground-to-aerial view synthesis and present two postulates. A common strategy adopted for text-based single image translation is to use a robust text-to-image pretrained model in a two-stage process. The first step finds the `optimized text embedding' $e_{opt}$ (in the vicinity of $e_{tgt}$) that best generates the `source' image $I_{S}$ and subsequently finetune the diffusion model to generate the `source' image $I_{S}$ using $e_{opt}$. In the second step, a linear interpolation of $e_{tgt}$ and $e_{opt}$ are used to generate the edited image $I_{T}$ from the finetuned neural network, i.e., the backward diffusion process is
\begin{equation}\label{equ:interpolation}
\resizebox{0.9\hsize}{!}{$
    x_{t-1}=x_t - f(x_t, t, \alpha e_{tgt}+(1-\alpha) e_{opt}),~t=T,\cdots, 0. $}
\end{equation}

A text-based single image translation approach~\cite{kawar2022imagic,hertz2022prompt,brooks2022instructpix2pix,kim2022diffusionclip} for ground-to-aerial generation overcomes multiple limitations in terms of data availability and generalization. However, the challenges involved in ground-to-aerial translation inhibit the direct application of existing text-based single image translation methods for ground-to-aerial generation. We present two postulates for text-based single-image translation in the context of ground-to-aerial generation.

\begin{prop}
Domain gap between the finetuning task (e.g., ground view generation) and target task (aerial view generation) hinders the diffusion model from generating accurate target views and introduces bias towards the source view.  
\end{prop}

Diffusion models are probabilistic models. They are trained~\cite{ho2020denoising} by optimizing the negative log-likelihood of the model distribution under the expectation of the data distribution. Further simplification of the equation for formulating the training loss function involves variance reduction. In the first step of finetuning, the diffusion model is being trained to reproduce the source image given the optimized text embedding, irrespective of the input random noise. Hence, it has a natural bias towards the source image. 

When the image space corresponding to the target text embedding is in the vicinity of the image space corresponding to the optimized text embedding, consistent with the variance within which the neural network was trained to generate, the generated target image is a high fidelity image consistent with the target text. When the desired transformation is large (ground-to-aerial), outside the limits of the variance, the diffusion model is unable to generate an aerial image. 

\begin{figure*}
    \centering
    \includegraphics[scale=0.42]{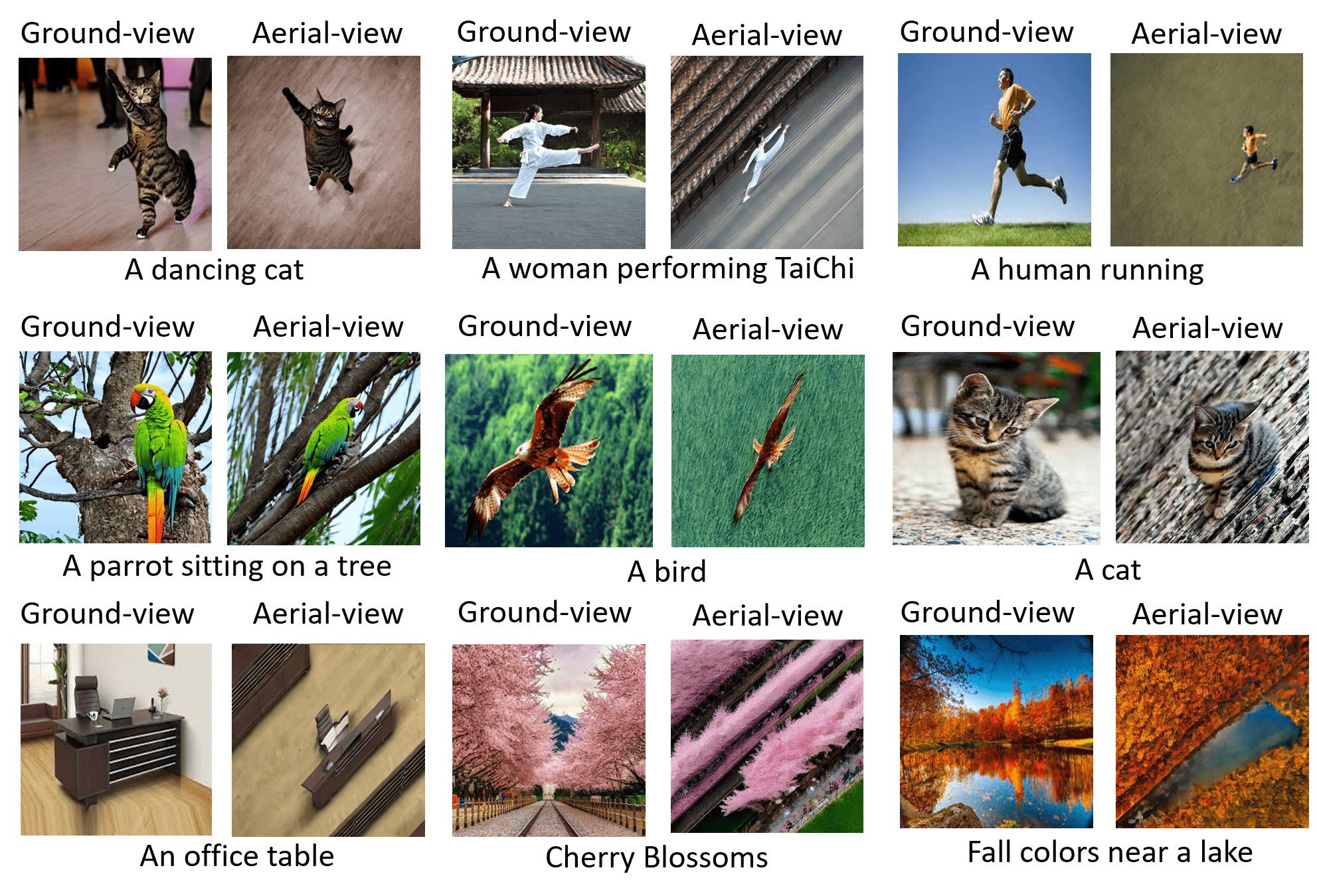}
    \caption{\textbf{Results.} We apply Aerial Diffusion on diverse images such as animals/ birds, natural scenes, human actions, indoor settings, etc and show that our method is able to generate high-quality high-fidelity aerial images.}
    \label{fig:supp_results1}
\end{figure*}

\begin{prop}
A finetuned diffusion model cannot generalize well to the target prompt if the text embedding and image spaces corresponding to the source and the target are very different and far away from each other on the nonlinear text-image embedding manifold. 
\label{sec:prop2}
\end{prop}

The embedding space and the corresponding image representation space are locally linear. Hence, when the target text embedding dictates a relatively small change to the source image, a linear combination between the optimized text embedding and the target text embedding generates a high-fidelity target image, faithful to the target text. 
% The sentences below are redundant.  Basically, when you have target and source images that involve large perspective changes (i.e. use of large rotation of camera poses), then the image representation space is no longer "locally linear".
% Moreover, since the representation spaces are locally linear, when we start moving in the direction of the target embedding from the optimized text embedding, the generated image gradually changes from the source image to a high-fidelity target image. Naturally, after a point, the fidelity of the generated image decreases. 
In contrast, when a linear interpolation of the text prompts in Eq.~\eqref{equ:interpolation} is applied 
% between the text embeddings corresponding 
to ground-to-aerial translation, depending on $\alpha$, the images generated are either high fidelity (but low target text faithfulness) or high target text faithfulness (but low fidelity). Moreover, the ground-view image doesn't gradually change to an oblique-view image followed by aerial-view image, the manifold is not smooth. Rather, the change is quite drastic and it is difficult to find an optimal solution in the linear interpolation space. 
Essentially, when there is a large perspective changes from the source to target images (i.e. involving large rotation of camera poses), the image representation space is no longer
``locally linear'', thereby linear interpolation is no longer adequate to generate high-fidelity images.
%\begin{figure}
%    \centering
%    \includegraphics[scale=0.28]{Figures/altsampling.png}
%    \caption{Linear interpolation~\cite{kawar2022imagic,zhang2022sine} between $e_{opt}$ and $e_{tgt}$ is unable to find a good solution because the representation spaces are not globally linear, while desired transformation for ground-to-aerial is large. We propose to use alternating text embeddings while sampling to find the optimal solution in a higher dimensional non-linear space. The embeddings shown in this 2D figure are for representation purposes only, the actual embeddings are in a high dimensional space.}
%    \label{fig:altsampling}
%\end{figure}

\subsection{Method}
\label{sec:method}

Motivated by the challenges described above, we propose Aerial Diffusion for text guided single-image ground-to-aerial translation. An overview of our solution is as follows. We start with a pretrained robust stable diffusion~\cite{rombach2022high} model as the backbone. Our method has three stages. In the first step, we preprocess the ground-view image $I_{S}$ with a carefully crafted homography transformation to generate $I_{Sh}$. This reduces the bias in the finetuning step. In the second step, we finetune the diffusion model by first optimizing the text-embedding within the vicinity of $e_{src}$ to find $e_{opt}$ that best generates $I_{Sh}$. Subsequently, we finetune the diffusion model to reconstruct $I_{Sh}$, given $e_{opt}$. In the third step on inferencing/sampling, we use an alternating strategy to manipulate the text embedding layer to generate a high-fidelity aerial image $I_{T}$. Next we describe each step in detail. 

\vspace{-0.5em}
\paragraph{Step 1: Preprocessing using a homography transformation.} The bias acquired by the diffusion model during the second step of finetuning inhibits large transformations. One way to decrease the bias is to reduce the number of iterations while finetuning. However, this leads to unsurprisingly low quality generated images. To decrease the bias while finetuning, we preprocess the ground-view image by transforming it with a 2D {\em homography transformation}~\cite{szeliski2022computer}  (inverse perspective mapping). This homography projects the ground-view image to its rough 2D projected aerial view. Note that we are unable to use a 3D homography mapping to obtain the 3D aerial view projection, a better pseudo estimate of the aerial view, due to the unavailability of camera matrix, multi-views, depth information, etc. On the other hand, depth estimation methods~\cite{fu2018deep,godard2017unsupervised} increase the complexity of the problem. 

Consider a 3D cube (Figure~\ref{fig:overview}). Without loss of generality, the 2D image captured by a ground-camera can be regarded as the projection of the scene in the front-face of the cube. A camera facing the top face of the cube will be able to capture the accurate 2D aerial view of the scene. Since we have no knowledge of the camera parameters corresponding to the ground-view image, we are unable to shift the camera to obtain a different view of the scene. With respect to the ground-camera, the 2D projection of the front-face of the cube on the bottom face of the cube is the best `aerial projection' that we can get (inverse perspective mapping~\cite{szeliski2022computer}). This aerial projection is nowhere close to the true aerial view and does not resemble the ground-view either. Hence, when the diffusion model is finetuned, the bias is much lower than what it would have been if the optimization/finetuning were done directly with the ground-view image. This is because of the disparities between the image space of $I_{Sh}$ and $e_{src}$/ $e_{tgt}$, ingrained in the pretrained network. Moreover, it provides a pseudo estimate of the direction in which the image needs to be transformed in order to generate its aerial view at the inference stage. 

We maintain the horizontal and vertical distance between the edges of the faces in the ground view and its projected aerial view, to better preserve the resolution and the aspect ratio. Formally, the coordinates (in order) of the corners of the ground-view image and the homography projected image are $\{(0,0), (H,0), (H,W), (0,W)\}$ and $\{(0,W), (H,0), (H,W), (0,2W)\}$ respectively (Figure~\ref{fig:overview}). The homography can then be computed and applied. 

\paragraph{Step 2: Finetuning the diffusion model.} We first optimize the text-embedding~\cite{kawar2022imagic,zhang2022sine} to generate $I_{Sh}$ and subsequently finetune the diffusion model using $e_{opt}$ to generate $I_{Sh}$. In contrast to popular text-based image editing approaches that find the optimized text embedding $e_{opt}$ in the vicinity of the target text embedding $e_{tgt}$, we find $e_{opt}$ in the vicinity of the source text embedding $e_{src}$. This is due to two reasons: (i) the disparity between the homography transformed view and the target text is still large (though much smaller than the disparity between the ground-view and target text). Hence, it is unlikely that a good $e_{opt}$ will be obtained when the optimization is run (around $e_{tgt}$) for a limited number of iterations. (ii) we do not want the network to develop a bias towards the homography image as the `aerial view'. 

To find $e_{opt}$, we freeze the parameters of the generative diffusion model $f_{\theta}$ and optimize $e_{src}$ using the denoising diffusion objective~\cite{ho2020denoising}. This optimization is run for a small number of iterations, in order to remain close to $e_{src}$ for meaningful embedding space manipulation at inferencing. 
\begin{equation}
    \min_{e_{opt}} \sum_{t=T}^0 L(f(x_t, t, e_{opt};\theta), I_{Sh}),
\end{equation}
\abovedisplayskip=-2pt
where $f(x,t,y)$ is the $t$-th backward diffusion step, $\theta$ denotes the U-net parameters, and $L$ is the denoising diffusion objective. To enable $e_{opt}$ reconstruct the $I_{Sh}$ with high fidelity, we finetune the diffusion model, again using the denoising diffusion objective~\cite{ho2020denoising,saharia2022image,kawar2022imagic}:
\begin{equation}
    \min_{e_{theta}} \sum_{t=T}^0 L(f(x_t, t, e_{opt};\theta), I_{Sh}).
\end{equation}

\begin{figure*}
\vspace*{-1em}
    \centering
    \includegraphics[scale=0.48]{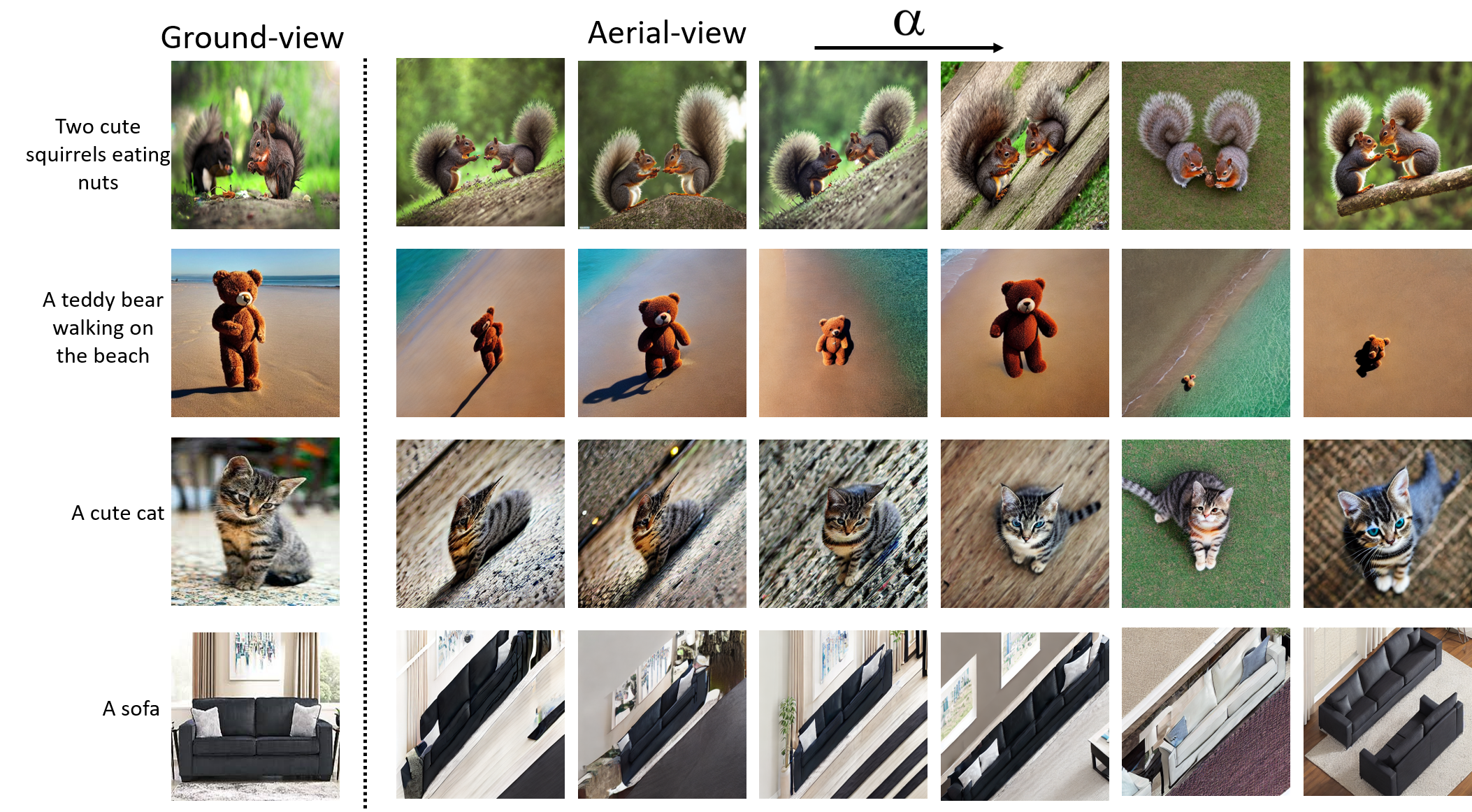}
    \caption{\textbf{Effect of $\alpha$.} Low values of $\alpha$ generate images that are less aerial, high values of $\alpha$ generate low-fidelity images. A trade-off between the viewpoint and fidelity generates high-fidelity aerial images. The transformation, with $\alpha$, is not smooth, reinforcing Postulate 2.}
    \label{fig:results2}
\end{figure*}

\paragraph{Step 3: Inferencing/ sampling by text embedding manipulation.} Our next step is to use the finetuned diffusion model to generate a high-fidelity aerial image. Prior work~\cite{kawar2022imagic,zhang2022sine} use linear interpolation between the optimized text embedding $e_{opt}$ and the target text embedding $e_{tgt}$. As described in Section~\ref{sec:theory}, linear interpolation is not the best solution for large transformations such as ground-to-aerial generation and is unable to generate high-fidelity aerial images. 

Sampling from stable diffusion~\cite{rombach2022high} involves iteratively denoising the image for $T$ steps conditioned by text, starting with random noise. To deal with the aforementioned issues, we propose to alternate between two text embeddings $e_{1}$ and $e_{2}$, starting with $e_{1}$. We designate $e_{1}$ as the target text embedding $e_{tgt}$. This imposes a strong constraint on the diffusion model to generate an aerial view image corresponding to the text description. The bias of the diffusion neural network motivates the network to generate an image whose details are close to the ground-view image. However, merely relying on the bias of the neural network to capture all details of the scene is severely insufficient. Hence, we designate $e_{2}$ to be the linear interpolation of $e_{opt}$ and $e_{tgt}$, controlled by the hyperparameter $\alpha$. The linear interpolation can be mathematically represented as $e_{2} = \alpha * e_{tgt} + (1-\alpha) * e_{opt}$. $e_{2}$ enables the network to generate a high fidelity image while retaining the aerial viewpoint. For very low values of $\alpha$, the generated image is less aerial, despite reinforcing the viewpoint to be aerial by applying $e_{1}$ alternatingly. This is because of the bias of the neural network. Very high values of $\alpha$ result in low fidelity images, some details of the generated aerial image are not consistent with the ground-view image. An optimal solution is by tuning $\alpha$.

Linear interpolation enforces the generation of an image consistent with a text embedding in the linear space between $e_{opt}$ and $e_{tgt}$. This is a reasonable when the desired change is small: when the image spaces corresponding to $e_{opt}$ and $e_{tgt}$ are closeby, linear interpolation works due to local linearity. When the desired change is large (such as ground-to-aerial translation), the image spaces corresponding to $e_{opt}$ and $e_{tgt}$ are not nearby. Since the representation spaces are not globally linear, it becomes essential to search for the solution in a much higher dimensional non-linear space. This is achieved by our alternating strategy. The pseudo code for the alternating strategy is given below. 
\begin{algorithm}
\caption{Alternate Prompting in backward diffusion enables the diffusion model generate a high-fidelity aerial image}
\label{alg:ap}
$x_T\sim\mathcal N(0,I)$\;
\For{$t\gets T$ \KwTo $0$}{
    \eIf{$t\% 2=0$}
    {
        $x_{t-1}=x_t - f(x_t, t, \alpha e_{tgt}+(1-\alpha) e_{opt}).$\;
    }{
        $x_{t-1}=x_t - f(x_t, t, e_{tgt}).$\;
    }
}
\end{algorithm}

While the sampling repetitively alternates between $e_{1}$ and $e_{2}$, it is more beneficial to use $e_{1}$ (over $e_{2}$) at the first iteration. When the diffusion process starts with $e_{1}$, the network generates starts by generating an aerial image with details weakly dictated by its bias. Subsequent iterations that alternate between $e_{2}$ and $e_{1}$ fortify the generation of a high fidelity aerial image. On the contrary, when the diffusion process starts with $e_{2}$, the generated image in the first iteration is less aerial though with very high fidelity. The bias, along with $e_{opt}$ serve as a strong prior towards a non-aerial viewpoint. Subsequent iterations that use $e_{1}$ are unable to overcome this strong prior to alter the viewpoint to aerial view. Hence, we start inferencing with $e_{1}$. 

\begin{figure}
    \centering
    \includegraphics[scale=0.25]{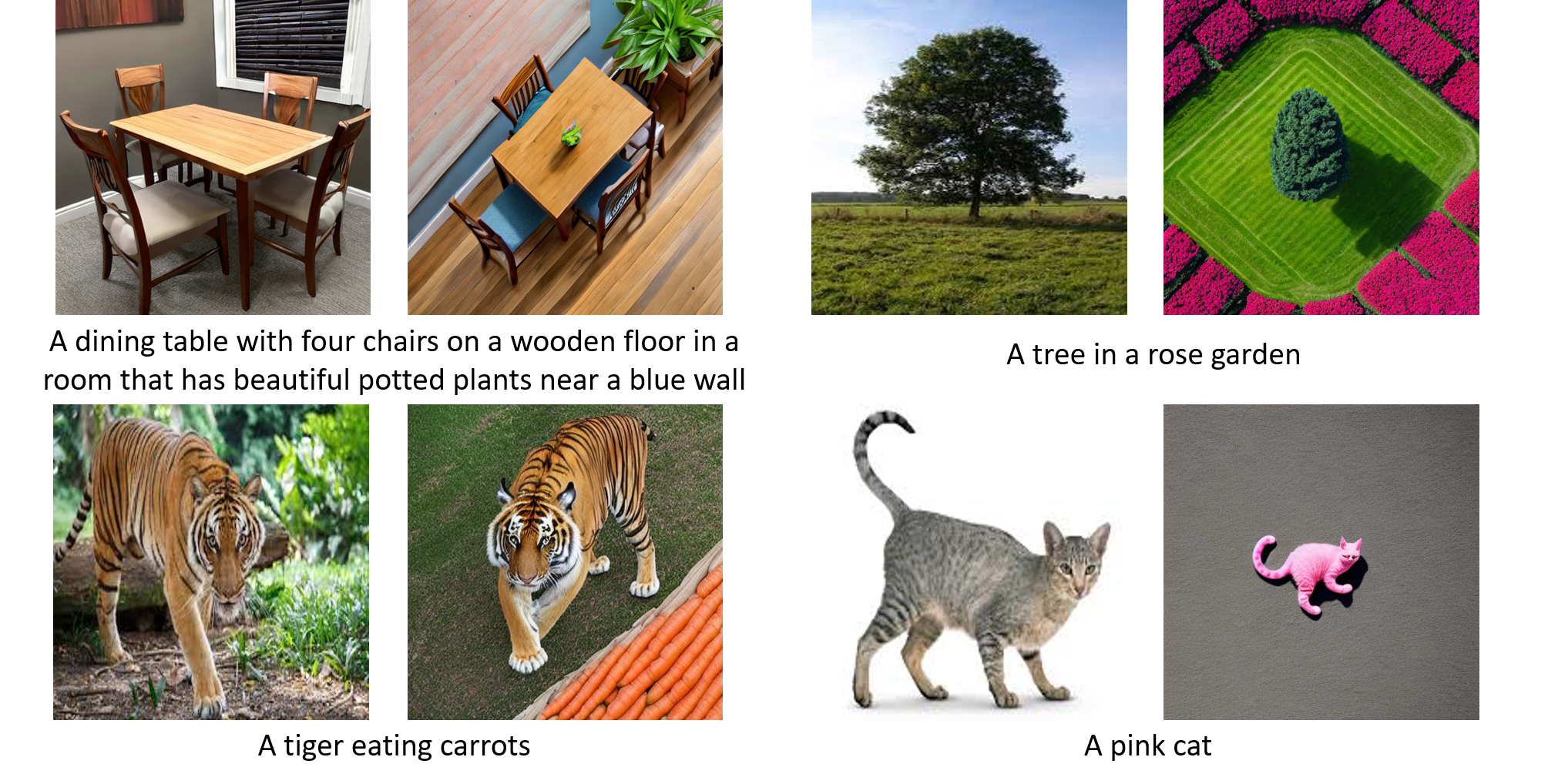}
    \caption{Based on the text description, Aerial Diffusion can generate aerial views with scene entities slightly different from the ground-view. The hallucination of background and unseen parts of the scene can also be controlled by text.}
    \label{fig:applications1}
\end{figure}

\section{Experiments and Results}

\subsection{Implementation details}

We use the stable diffusion~\cite{rombach2022high} text-to-image model as the backbone architecture. It has been pretrained on the massive text-image LAION-5B dataset (laion2B-en, laion-high-resolution, laion-improved-aesthetics). Prior to the homography transformation, we resize all images to a resolution of $256 \times 256$. We use OpenCV to apply the homography transformation on the image to generate a homography transformed image of resolution $512 \times 512$, which is used to optimize the text embedding and the diffusion model in the next step. We finetune the text embedding for $500$ iterations with a learning rate of $1e-3$ using the Adam optimizer, and the diffusion model for $1000$ iterations at a learning rate of $2e-6$. For each image, this entire optimization takes between $8$ and $9$ minutes on one NVIDIA RTX A5000 GPU with 24GB memory.

While sampling/inferencing using the finetuned diffusion model, we iteratively refine the image, starting with random noise, for $T=50$ iterations. At every iteration, the diffusion model is applied on the refined image from the previous iteration, as per the standard procedure~\cite{ho2020denoising} in sampling from diffusion models. The text embedding condition, while sampling at each iteration, alternates between $e_{1}$ and $e_{2}$, starting from $e_{1}$. The guidance scale is set to $7.5$.

\begin{figure*}
    \vspace*{-1em}
    \centering
    \includegraphics[scale=0.4]{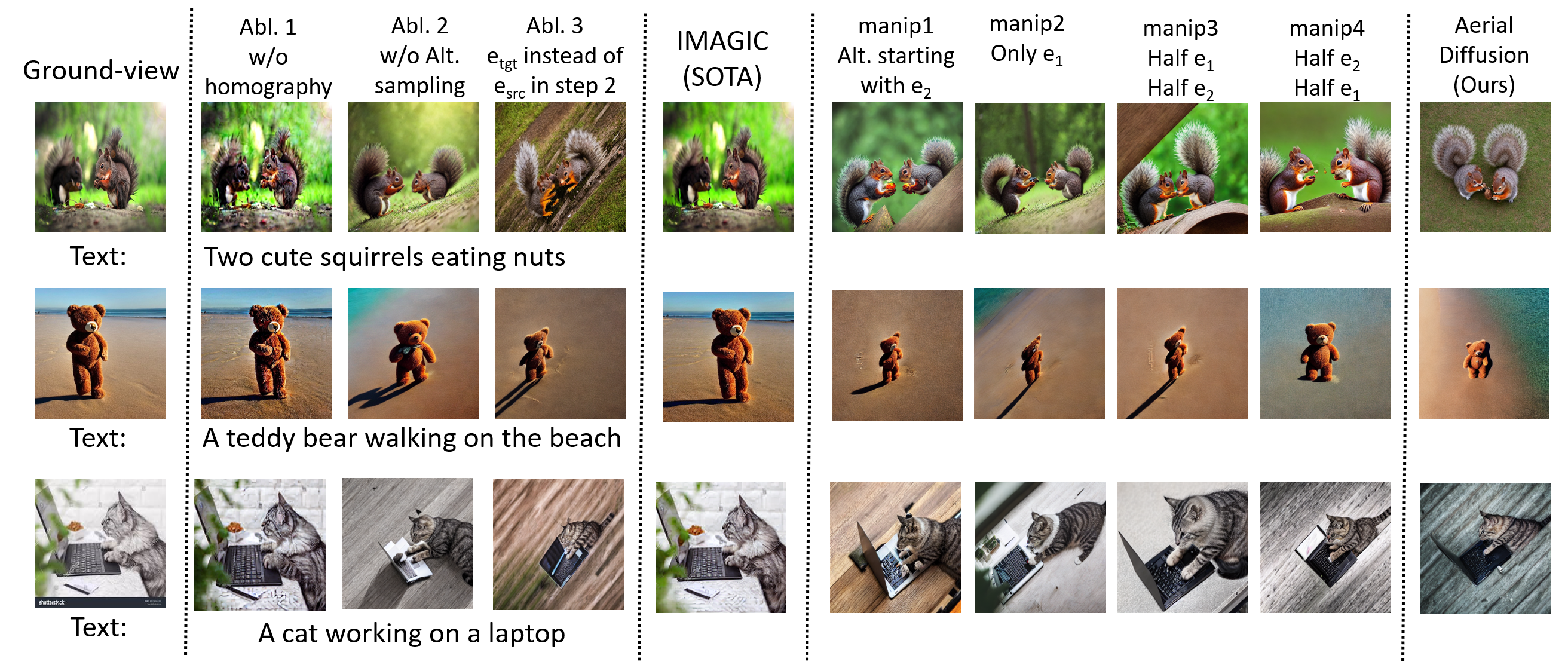}
    \caption{\textbf{Ablations and Comparisons.} Ablations: we prove the effectiveness of the homography, our alternating strategy over linear interpolation and finetuning with $e_{src}$ instead of $e_{tgt}$. SOTA Comparisons: IMAGIC ~\cite{kawar2022imagic} (CVPR 2023) is unable to generate aerial views due to high bias towards the input image, domain gap and restricting the solution search to the linear interpolation manifold. Comparisons with other embedding space manipulation strategies that utilize both $e_{1}$ and $e_{2}$ reveal that our Alternating strategy is better.}
    \label{fig:results3}
\end{figure*}

\begin{figure}
    \centering
    \includegraphics[scale=0.25]{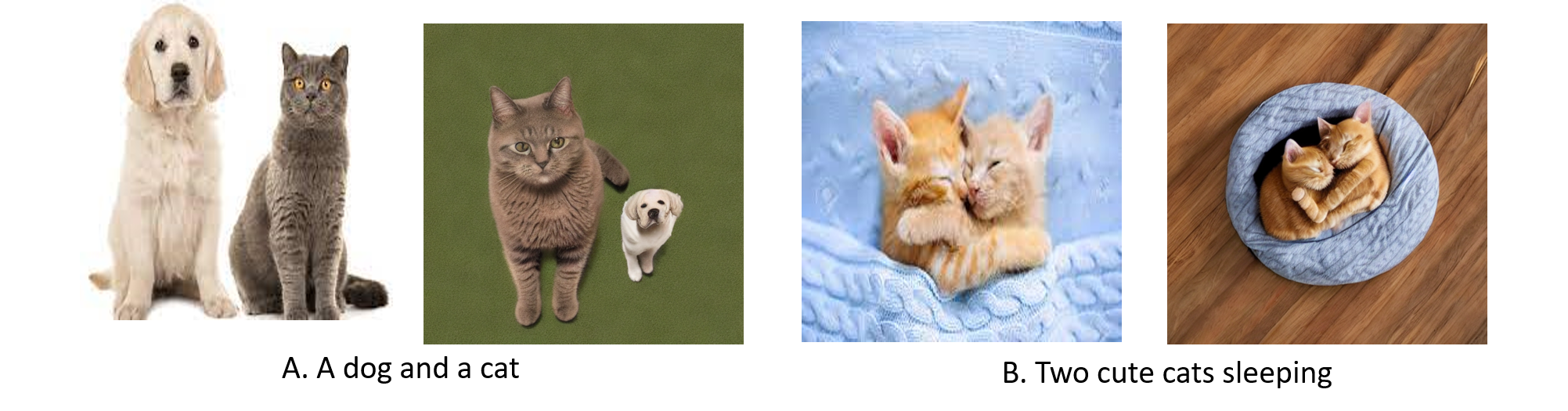}
    \caption{\textbf{Failure cases.} A. The identity of the dog and cat is interchanged. B. The fidelity of the background (bed) is lost.}
    \label{fig:fail1}
\end{figure}

\subsection{Qualitative evaluation}

We apply our method on a number of real images from various domains including nature, human actions, buildings, etc. We collect most of these images from Google Images/ Flickr. We show the results in Figure~\ref{fig:teaser} and Figure~\ref{fig:results1}. Aerial images can be captured from varying heights and angles (including side/oblique views). We do not constrain the network on the height/ angle, hence, the diffusion model generates an aerial image of arbitrary height/ angle dictated by the random noise. For each image, we choose the best result corresponding to $0 < \alpha < 1$. Aerial Diffusion is successful in generating the aerial view, given its ground-view. It is able to hallucinate the aerial view of the entities in the scene (encompassing unseen aspects) as well as the background; while being faithful to the details in the ground-view. Since the underlying diffusion model is probabilistic, we get different results for different random seeds, all the generated images are faithful to the details in the ground-view as well as conform to the viewpoint being aerial. This diversity allows users to choose among a variety of options and is also a useful property in curating synthetic datasets.  

\noindent \textbf{More applications.} We show in Figure~\ref{fig:applications1} that our method is capable of generating aerial views consistent with the text description, even when the text dictates an aerial view with scene entities slightly different from the ground-view. Moreover, background hallucination can be controlled by the text description as well. 

\noindent \textbf{Effect of $\alpha$.} We show the effect of $alpha$ in Figure~\ref{fig:results2}. The finetuned diffusion model has high bias towards the ground-view image. Hence, the value of $\alpha$ needs to be carefully tuned in order to generate a high fidelity aerial image. A low value of $\alpha$ implies higher weight to the optimized text embedding and low weight to `aerial view'. This leads to the generated image being very similar to the homography image, the viewpoint of the generated image is less aerial. A high value of $\alpha$ implies higher weight to `aerial view'. The generated image is an aerial image, though with low fidelity. While the details in the generated aerial image are not completely different from the details in the ground-view image, due to the bias of the finetuned diffusion model, the fidelity or conformance to details contained in the ground-view image is still low. A good trade-off between fidelity w.r.t. ground-view image and `aerial view' is achieved at mid-values of $\alpha$. The change from ground-view to aerial-view as $\alpha$ varies from $0$ to $1$ is not gradual, reinforcing Postulate 2. 

\subsection{Ablations and comparisons}

\noindent \textbf{Ablating the model.} We show ablation experiments in Figure~\ref{fig:results3}. Ablation 1 uses the ground-view image instead of the homography transformed image to finetune the diffusion model. It proves that the homography transformation is successful in significantly lowering the bias of the model and helps generate aerial views, reinforcing our solution to Postulate 1. Ablation 2 uses linear interpolation for sampling instead of our alternating strategy. Results with the alternating strategy are better (high fidelity images with faithfulness to `aerial view') than the results with linear interpolation, justifying our solution to Postulate 2. Ablation 3 finds $e_{opt}$ in the vicinity of $e_{tgt}$ instead of $e_{src}$. Generated images are more aerial when $e_{opt}$ is optimized in the vicinity of $e_{src}$, proving that it helps in reducing the bias. 

\begin{figure*}
    \centering
    \includegraphics[scale=0.42]{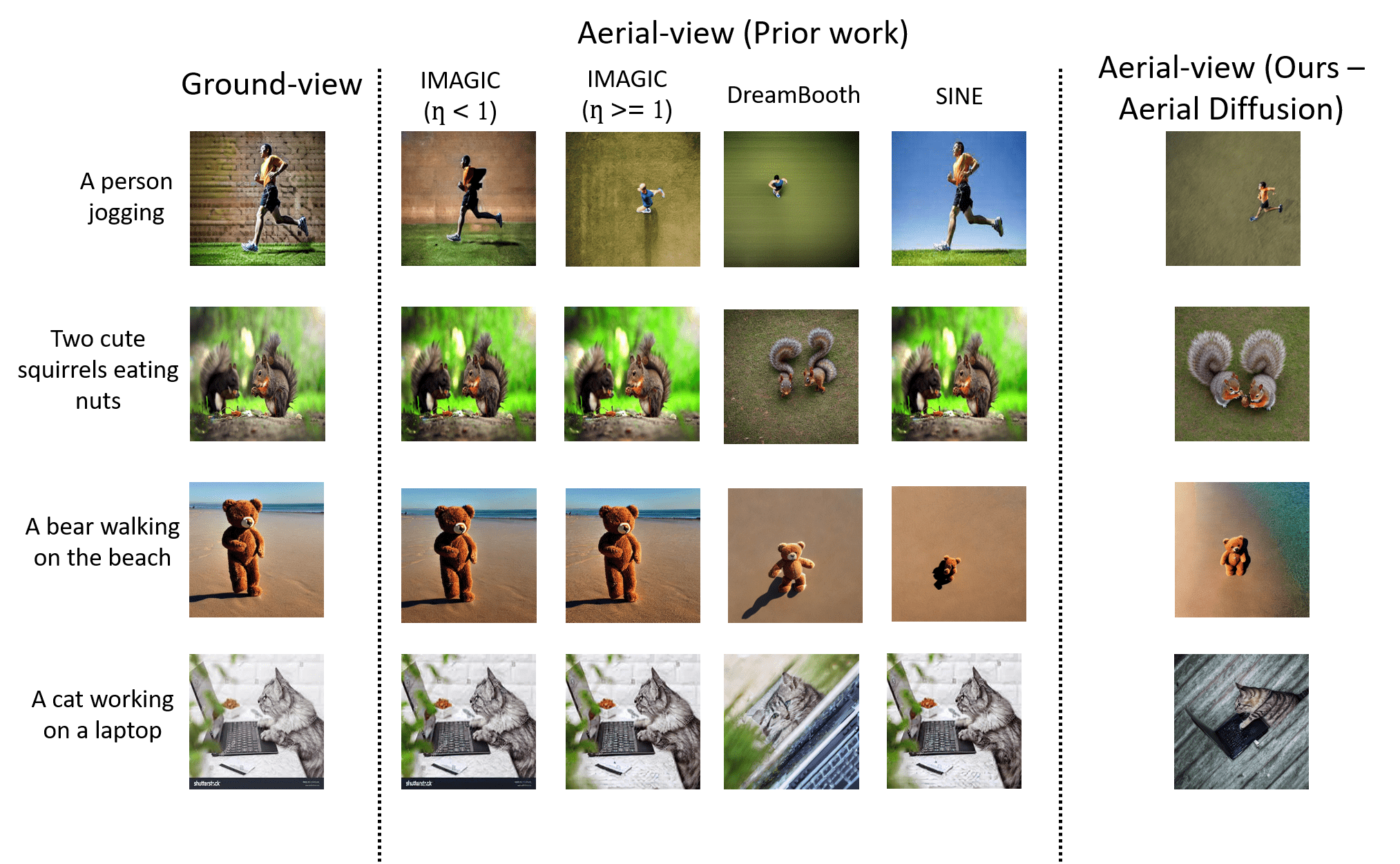}
    \caption{\textbf{State-of-the-art comparisons.} We compare with the state-of-the-art text-based image translation method, IMAGIC~\cite{kawar2022imagic} (CVPR 2023). IMAGIC is unable to generate a high-fidelity {\bf aerial-view} image. It merely reproduces the ground-view image, despite tuning all hyperparameters exhaustively. The hyperparameter $\eta$ (in IMAGIC~\cite{kawar2022imagic}) takes values between $0$ and $1$, and is unable to generate a high fidelity {\em aerial} image for any value of $\eta$. When $\eta$ is increased to a value higher than $1$, the model generates an aerial image but the fidelity with respect to the ground-view is completely lost. Other methods (contemporary/prior to IMAGIC~\cite{kawar2022imagic} (CVPR 2023)) such as DDIB~\cite{su2022dual} (ICLR 2023), DreamBooth~\cite{ruiz2022dreambooth} (CVPR 2023), SINE~\cite{zhang2022sine} (CVPR 2023), SDEdit~\cite{meng2021sdedit}(ICLR 2022) and Text2LIVE~\cite{bar2022text2live}(ECCV 2022) face the same issues. These are due to the challenges involved in ground-to-aerial translation described in Section 1 and Section 3.1 of our paper.}
    \label{fig:supp_results6}
\end{figure*}

\noindent \textbf{SOTA Comparisons.} We compare with IMAGIC~\cite{kawar2022imagic} (CVPR 2023), DreamBooth~\cite{ruiz2022dreambooth} (CVPR 2023), SINE~\cite{zhang2022sine}, SOTA text-based single image translation methods, in Figure~\ref{fig:results3}, \ref{fig:supp_results6}. Our method is far superior than prior art for ground-to-aerial translation, which are is unable to effectively perform ground-to-aerial translation due the high bias incurred while finetuning and searching for the solution in a limited linear interpolation space. 

\noindent \textbf{Comparisons with other text embedding manipulation methods.} We compare with other strategies to manipulate the text embedding space using $e_{1}$ and $e_{2}$ in Figure~\ref{fig:results3}. In manip1, we condition on $e_{2}$ and $e_{1}$ alternatingly, starting from $e_{1}$. Clearly, it is more beneficial to start sampling from $e_{1}$ as explained in Section~\ref{sec:method}. In manip2, we use just $e_{1}$ (text embedding corresponding to aerial view) to sample and rely on the bias of the network to generate the aerial image. Our alternating sampling method is able to generate higher fidelity aerial images. In manip3, for the first $T/2$ iterations, we sample using $e_{1}$ and for the second $T/2$ iterations, we sample using $e_{2}$. In manip4, for the first $T/2$ iterations, we sample using $e_{2}$ and for the second $T/2$ iterations, we sample using $e_{1}$. These experiments prove that our alternating sampling strategy works best. 

\noindent \textbf{Quantitative evaluation - user study.} Text guided single image ground-to-aerial translation is a recent development, and Aerial Diffusion is the first solution towards this goal. As such, no standard benchmark (and ground-truth) or quantitative metrics exist for evaluation. We evaluate Aerial Diffusion via human perceptual evaluation and observe that Aerial Diffusion is able to generate high-fidelity aerial images. We conduct the following types of evaluation:
\begin{enumerate}
    \item {\bf Image Quality: } Given a ground-view image and an aerial-view image generated using Aerial Diffusion, we ask participants to determine if the generated aerial-view image is a high-fidelity (w.r.t. ground-view image) aerial-view image; and rate the image on the 5-point Likert scale. The average rating over 10 images (rated by 49 participants) is 3.289.   
    
    \item {\bf Alternating Sampling: } Given a ground-view image and two aerial-view images generated using Aerial Diffusion and Ablation 2 (i.e. Aerial Diffusion without the Alternating Sampling method) respectively, we ask participants to choose the better high-fidelity aerial-view image. {\bf 83.05}$\%$ of the participants rate the image generated using Aerial Diffusion as the one with higher quality. 
    
    \item {\bf Reference for Aerial Diffusion:} Given a ground-view image and two aerial-view images generated using Aerial Diffusion and Ablation 3 (Aerial Diffusion with $e_{tgt}$ instead of $e_{src}$ while training) respectively, we ask participants to choose the higher fidelity aerial-view image. {\bf 78.125}$\%$ of the participants rate the image generated using Aerial Diffusion as the one with higher quality. 
   
\end{enumerate}

\noindent \textbf{Quantitative evaluation - metrics.} Our method uses prior knowledge (from robust pretraining), along with the knowledge gained while finetuning, to generate aerial images. It \textit{hallucinates} large parts and views of the scene that it has not encountered before. Hence, comparisons against other ground-to-aerial methods~\cite{regmi2019cross,toker2021coming,ding2020cross,shen2021cross}, that learn a specific data distribution by training on (an entire dataset with) paired data (and auxiliary information such as depth, semantic maps) is not relevant to this paper. In line with prior work~\cite{nichol2021glide,zhang2022sine,hertz2022prompt,kawar2022imagic,ruiz2022dreambooth} on text-based image translation, we report two metrics: (i) LPIPS to evaluate fidelity of aerial image w.r.t. ground-view image - the average score for $1-$ LPIPS is 0.352 (higher, the better). However, similarity scores (such as FID, LPIPS) compute patch-wise image similarity and the aerial view is much
different from the ground-view. No better evaluation strategy is available and is a direction for future work. (ii) CLIP score to evaluate the alignment of the generated aerial image with the text (dictating the viewpoint to be aerial) - the average CLIP score is 0.3233 (higher, the better). For reference, text-based image-editing methods such as Imagic~\cite{kawar2022imagic} (CVPR 2023) and DreamBooth~\cite{ruiz2022dreambooth} (CVPR 2023) achieve CLIP scores of $0.25$ to $0.3$ on image editing dictating far lesser change than front-to-aerial translation.  

%\vspace*{0.25em}
\noindent \textbf{More results.} Please refer to the supplementary material. 
\section{Conclusion, Limitations and Future Work}

In this paper, we introduce a novel method, Aerial Diffusion, for generating aerial views from a single ground-view image using text as a guiding factor. We use homography as guidance and a diffusion model on the generated image with an alternating denoising direction based on switching between viewpoint text embedding and fidelity of the generated images. Our method has some limitations. Our method has some limitations: (i) the homography transformation results in a directional (diagonal) bias in the generated aerial image in many cases; (ii) it is limited to the knowledge contained in the pretrained stable diffusion model and is unable to hallucinate scenes~\cite{somepalli2022diffusion} outside of this domain; (iii) the value of $\alpha$ in sampling needs to be manually tuned; (iv) the problem domain of unpaired ground-to-aerial does not have concrete quantitative analysis metrics. Future work can focus on the development of methods that overcome these limitations. Other directions include extending Aerial Diffusion to complex scenes with multiple objects (and an intricate background), generating higher-fidelity images, extending the method to videos, using the synthetic aerial data for aerial video analysis, detection, and recognition tasks. 

\noindent \textbf{Acknowledgements:} This research has been supported by ARO Grants W911NF2110026 and Army Cooperative Agreement W911NF2120076   

\begin{figure*}
    \centering
    \includegraphics[scale=0.5]{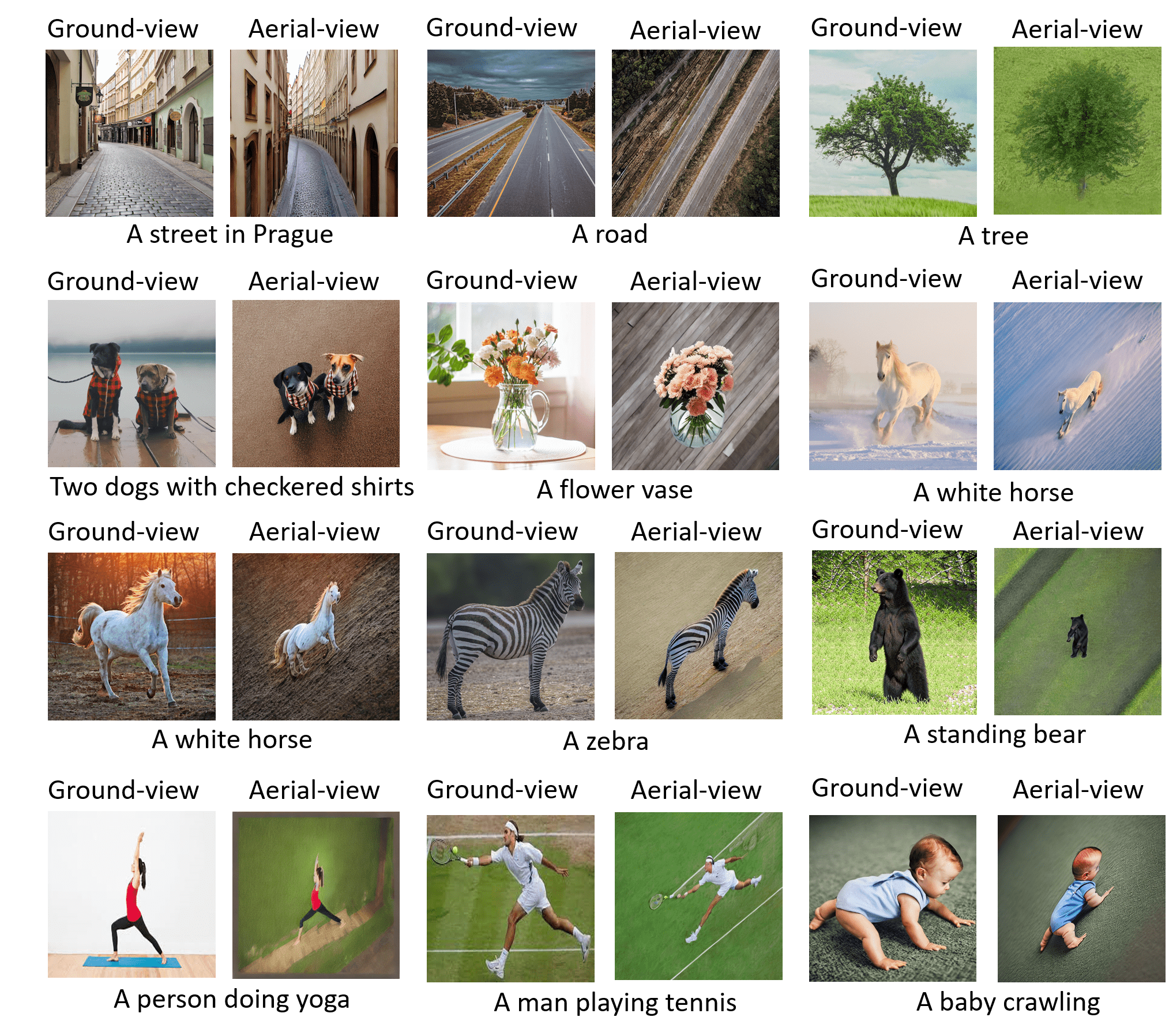}
    \caption{\textbf{Results.} We apply Aerial Diffusion on diverse images such as animals/ birds, natural scenes, human actions, indoor settings, etc and show that our method is able to generate high-quality high-fidelity aerial images.}
    \label{fig:supp_results2}
\end{figure*}

\begin{figure*}
    \centering
    \includegraphics[scale=0.5]{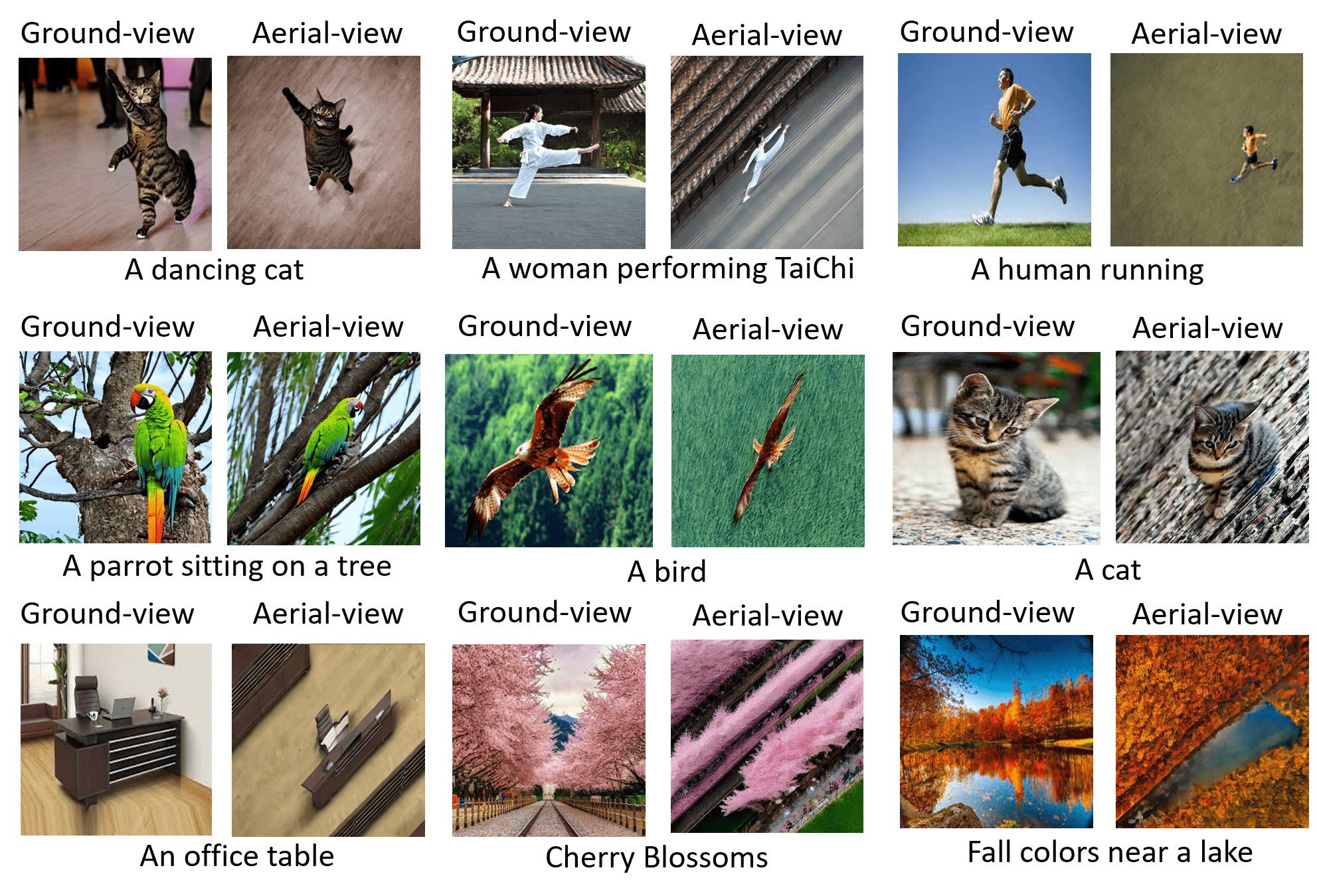}
    \caption{\textbf{Results.} We apply Aerial Diffusion on diverse images such as animals/ birds, natural scenes, human actions, indoor settings, etc and show that our method is able to generate high-quality high-fidelity aerial images.}
    \label{fig:supp_results3}
\end{figure*}

\begin{figure*}
    \centering
    \includegraphics[scale=0.45]{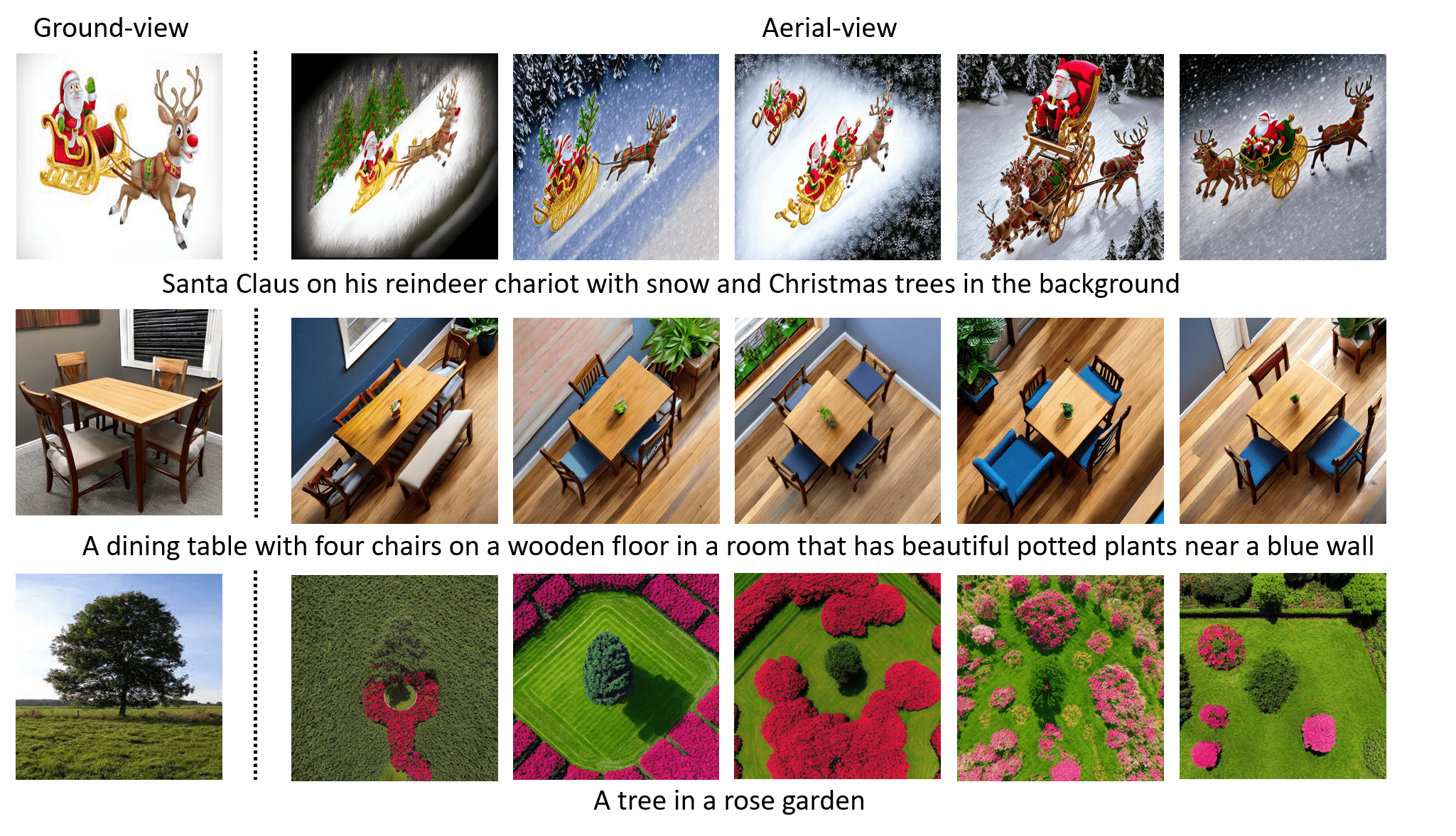}
    \caption{\textbf{Results.} Based on the text description, Aerial Diffusion can generate aerial views with scene entities slightly different from the ground-view. The hallucination of background and unseen parts of the scene can also be controlled by text.}
    \label{fig:supp_results4}
\end{figure*}

\begin{figure*}
    \centering
    \includegraphics[scale=0.5]{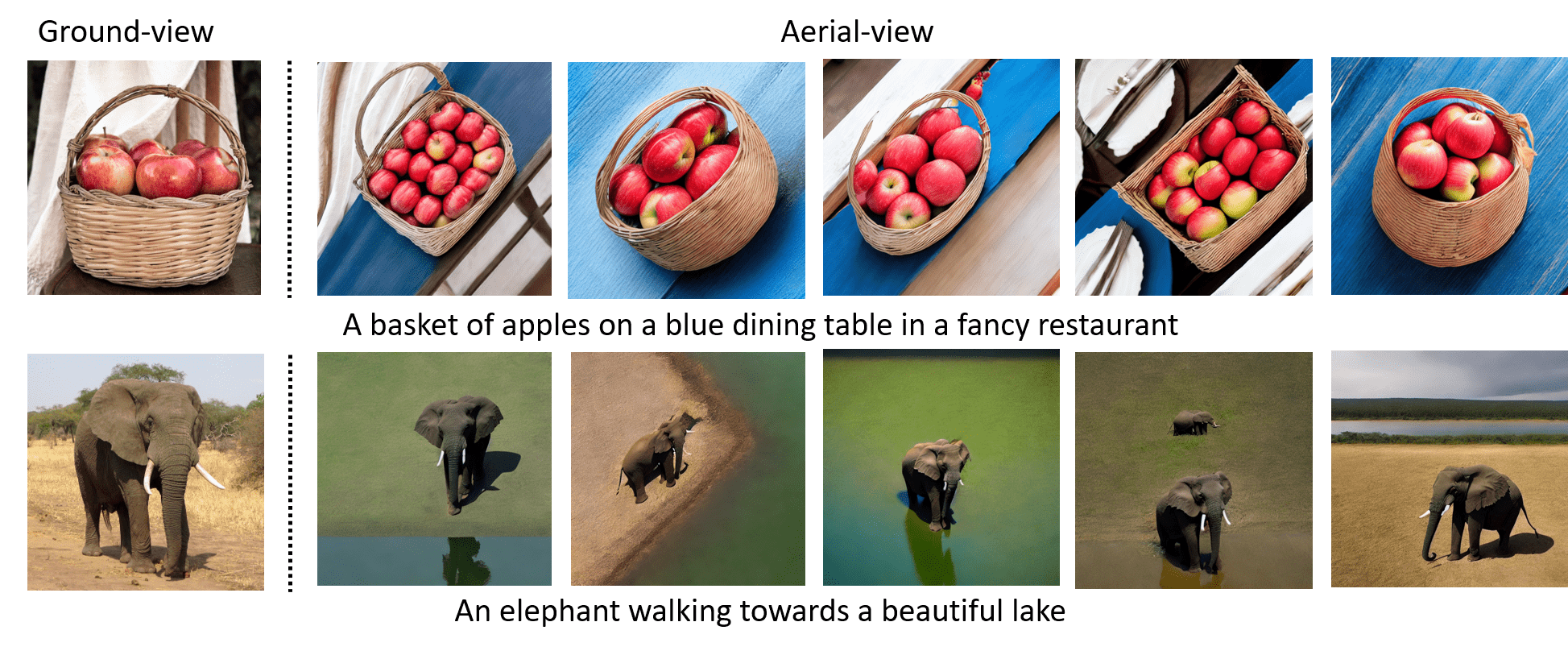}
    \caption{\textbf{Results.} Based on the text description, Aerial Diffusion can generate aerial views with scene entities slightly different from the ground-view. The hallucination of background and unseen parts of the scene can also be controlled by text.}
    \label{fig:supp_results4}
\end{figure*}

\begin{figure*}
    \centering
    \includegraphics[scale=0.5]{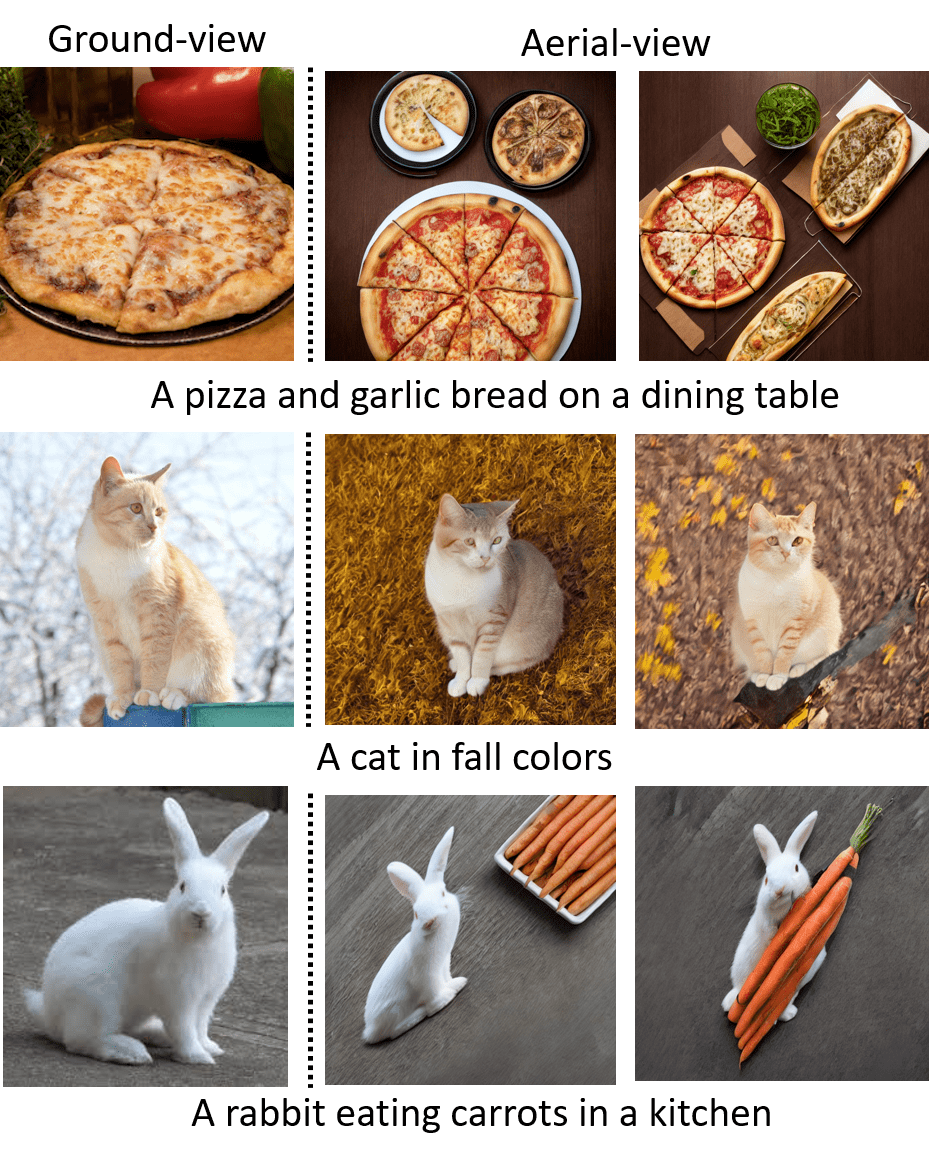}
    \caption{\textbf{Results.} Based on the text description, Aerial Diffusion can generate aerial views with scene entities slightly different from the ground-view. The hallucination of background and unseen parts of the scene can also be controlled by text.}
    \label{fig:supp_results4}
\end{figure*}

\begin{figure*}
    \centering
    \includegraphics[scale=0.48]{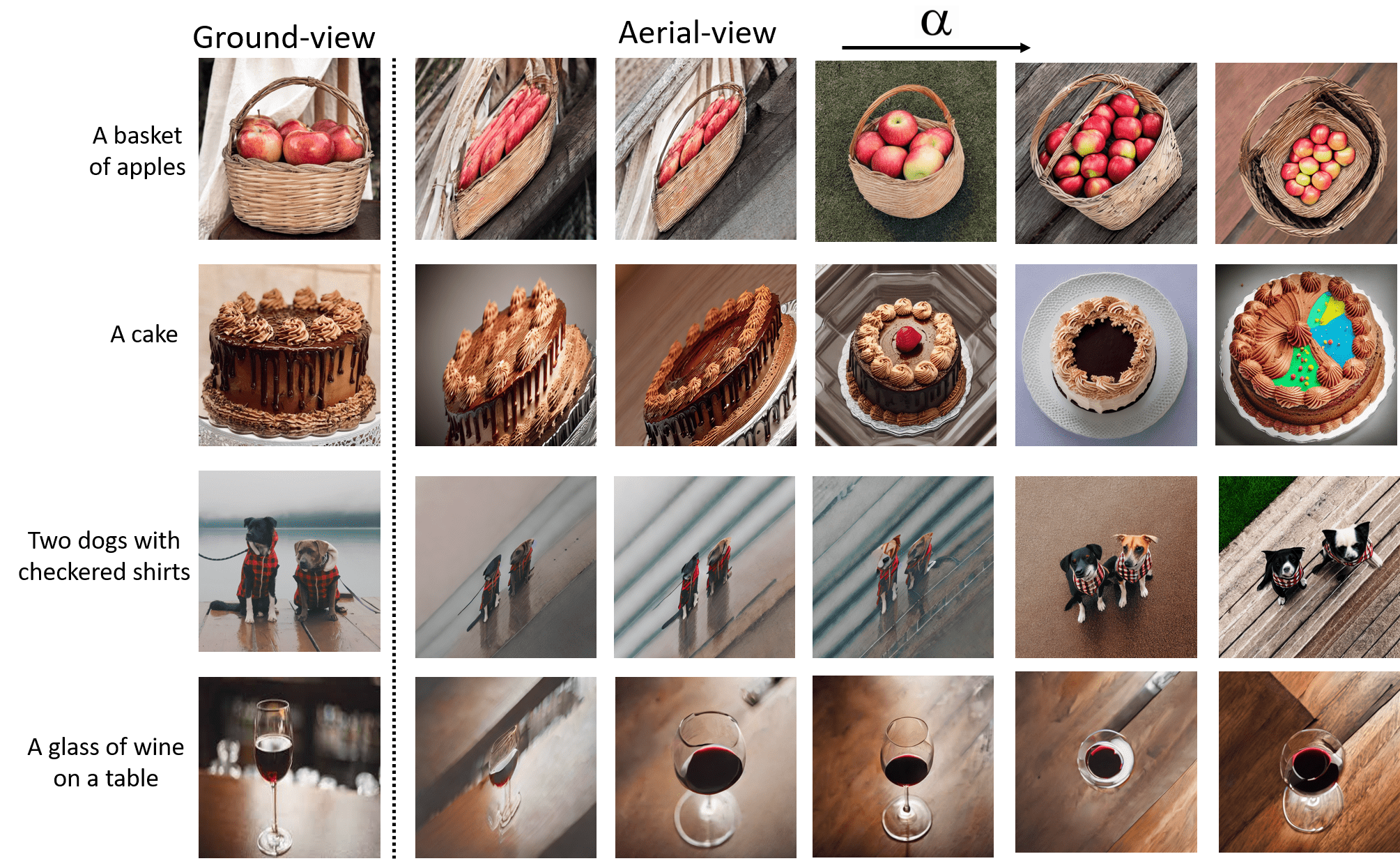}
    \caption{\textbf{Effect of $\alpha$.} Low values of $\alpha$ generate images that are less aerial, high values of $\alpha$ generate low-fidelity images. A trade-off between the viewpoint and fidelity generates high-fidelity aerial images. The transformation, with $\alpha$, is not smooth, reinforcing Postulate 2.}
    \label{fig:supp_results5}
\end{figure*}

\begin{figure*}
    \centering
    \includegraphics[scale=0.45]{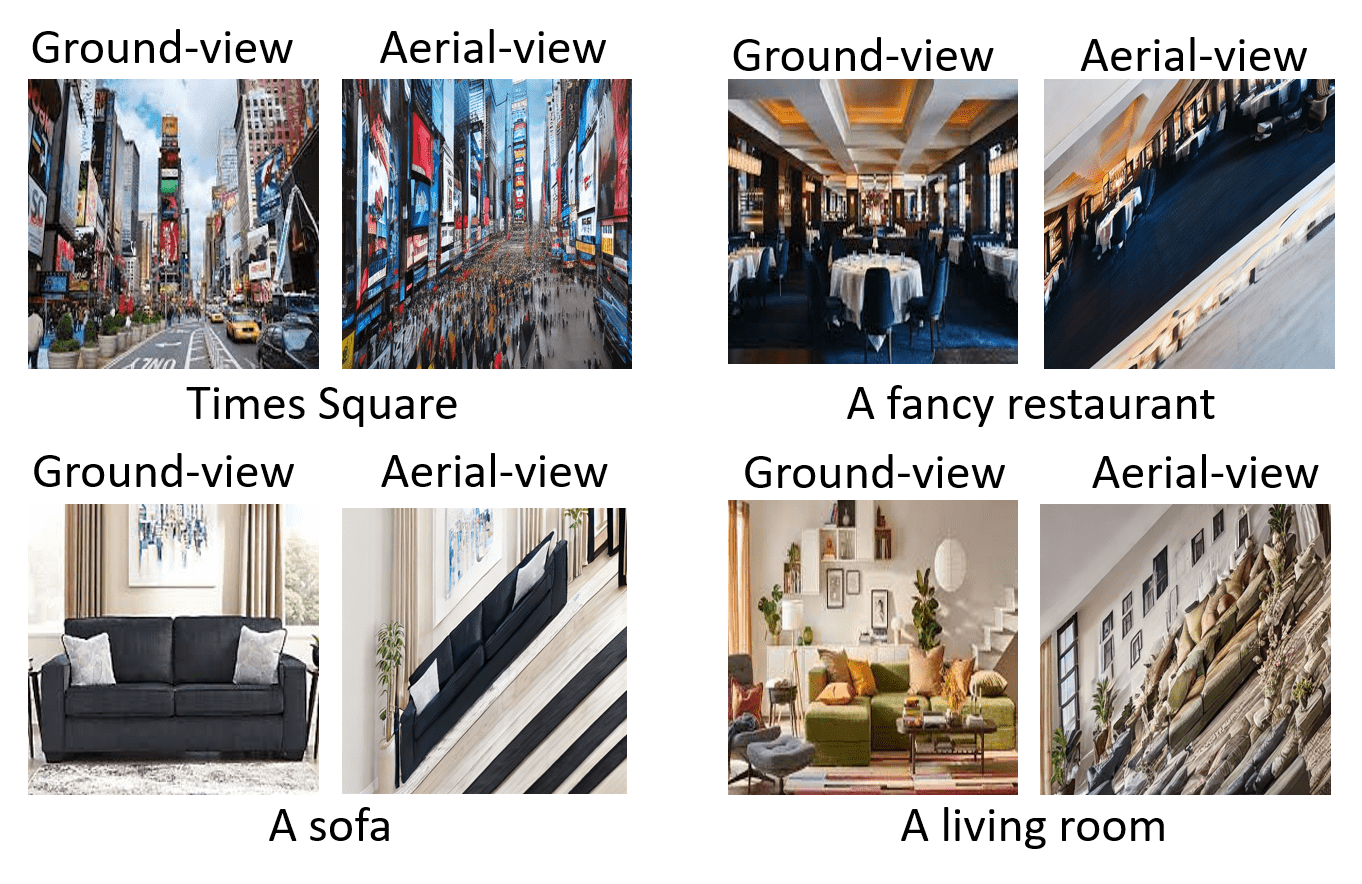}
    \caption{\textbf{Failure cases.} Our method is unable to generate high fidelity aerial views of scenes that have high complexity with multiple objects or scene entities, especially when the {\em text description} of the scene is {\em brief and inadequate}. This is a direction for future work.}
    \label{fig:supp_results7}
\end{figure*}

{\small
\bibliographystyle{ieee_fullname}
\bibliography{references}
}

\end{document}